\documentclass[10pt,twocolumn,letterpaper]{article}

\usepackage[pagenumbers]{cvpr} 

\usepackage{amsmath}
\usepackage{amssymb}
\usepackage{booktabs}
\usepackage{multirow}
\usepackage{amsfonts}
\usepackage{bbding}
\usepackage{amsmath}
\usepackage{array}
\usepackage{colortbl}
\usepackage{color}
\usepackage{makecell}
\usepackage{balance}
\usepackage[T1]{fontenc}
\usepackage{graphicx}
\usepackage{subfloat}
\usepackage{lipsum}
\usepackage[accsupp]{axessibility}  

\newcommand\blfootnote[1]{%
\begingroup
\renewcommand\thefootnote{}\footnote{#1}%
\addtocounter{footnote}{-1}%
\endgroup
}

\definecolor{my_blue}{rgb}{0.58,0.73,0.91}
\definecolor{my_red}{rgb}{0.91,0.63,0.58}

\definecolor{my_self}{RGB}{236, 242, 255}
\definecolor{my_sup}{RGB}{227, 223, 253}
\definecolor{my_semi}{RGB}{255, 244, 210}

\usepackage{enumitem}
\def\cam{\textcolor{black}}
\usepackage[pagebackref,breaklinks,colorlinks]{hyperref}

\usepackage[capitalize]{cleveref}
\crefname{section}{Sec.}{Secs.}
\Crefname{section}{Section}{Sections}
\Crefname{table}{Table}{Tables}
\crefname{table}{Tab.}{Tabs.}

\def\cvprPaperID{***} 
\def\confName{CVPR}
\def\confYear{2023}

\begin{document}

\title{Learning to Fuse Monocular and Multi-view Cues for \\ Multi-frame Depth Estimation in Dynamic Scenes}

\author{Rui Li\textsuperscript{1}, Dong Gong\textsuperscript{2*}, Wei Yin\textsuperscript{3*}, Hao Chen\textsuperscript{4}, Yu Zhu\textsuperscript{1}, Kaixuan Wang\textsuperscript{3}, \\  Xiaozhi Chen\textsuperscript{3}, Jinqiu Sun\textsuperscript{1}, Yanning Zhang\textsuperscript{1*}\\
\textsuperscript{1}Northwestern Polytechnical University,  \textsuperscript{2}The University of New South Wales, \textsuperscript{3}DJI, \textsuperscript{4}Zhejiang University\\
{\tt \small \href{https://github.com/ruili3/dynamic-multiframe-depth}{https://github.com/ruili3/dynamic-multiframe-depth}}
}

\maketitle
\begin{abstract}
Multi-frame depth estimation generally achieves high accuracy relying on the multi-view geometric consistency. When applied in dynamic scenes, e.g., autonomous driving, this consistency is usually violated in the dynamic areas, leading to corrupted estimations. Many multi-frame methods handle dynamic areas by identifying them with explicit masks and compensating the multi-view cues with monocular cues represented as local monocular depth or features. The improvements are limited due to the uncontrolled quality of the masks and the underutilized benefits of the fusion of the two types of cues. In this paper, we propose a novel method to learn to fuse the multi-view and monocular cues encoded as volumes without needing the heuristically crafted masks. As unveiled in our analyses, the multi-view cues capture more accurate geometric information in static areas, and the monocular cues capture more useful contexts in dynamic areas. To let the geometric perception learned from multi-view cues in static areas propagate to the monocular representation in dynamic areas and let monocular cues enhance the representation of multi-view cost volume, we propose a cross-cue fusion (CCF) module, which includes the cross-cue attention (CCA) to encode the spatially non-local relative intra-relations from each source to enhance the representation of the other. Experiments on real-world datasets prove the significant effectiveness and generalization ability of the proposed method.

\end{abstract}

\section{Introduction}

\label{sec:intro}
\blfootnote{\textsuperscript{*}Corresponding author}
Depth estimation is a fundamental and challenging task for 3D scene understanding in various application scenarios, such as autonomous driving \cite{geiger2012we,li2022bevdepth,guizilini20203d}. 
With the advent of convolutional neural networks (CNNs) \cite{he2016deep, gong2017motion}, depth estimation methods \cite{yao2018mvsnet,bhat2021adabins,yuan2022new,li2020enhancing,li2023learning,li2022self,yin2021learning} are capable of predicting promising results given either single or multiple images. The single image-based methods learn the monocular cues, \eg, the texture or object-level features, to predict the depth \cite{yin2021virtual,yin2022towards,bhat2021adabins}
, while multi-frame methods \cite{yao2018mvsnet,watson2021temporal,wimbauer2021monorec} can generally obtain higher overall accuracy relying on the multi-view geometric cues.
Specifically, the 3D cost volume has been proven simple and effective for depth estimation, which encodes the multi-frame matching probabilities with a set of depth hypotheses \cite{yao2018mvsnet,wimbauer2021monorec,gu2020cascade}. 
\par

\begin{figure}[t]
\centering
\vspace{-12pt}
\includegraphics[width=0.95\linewidth]{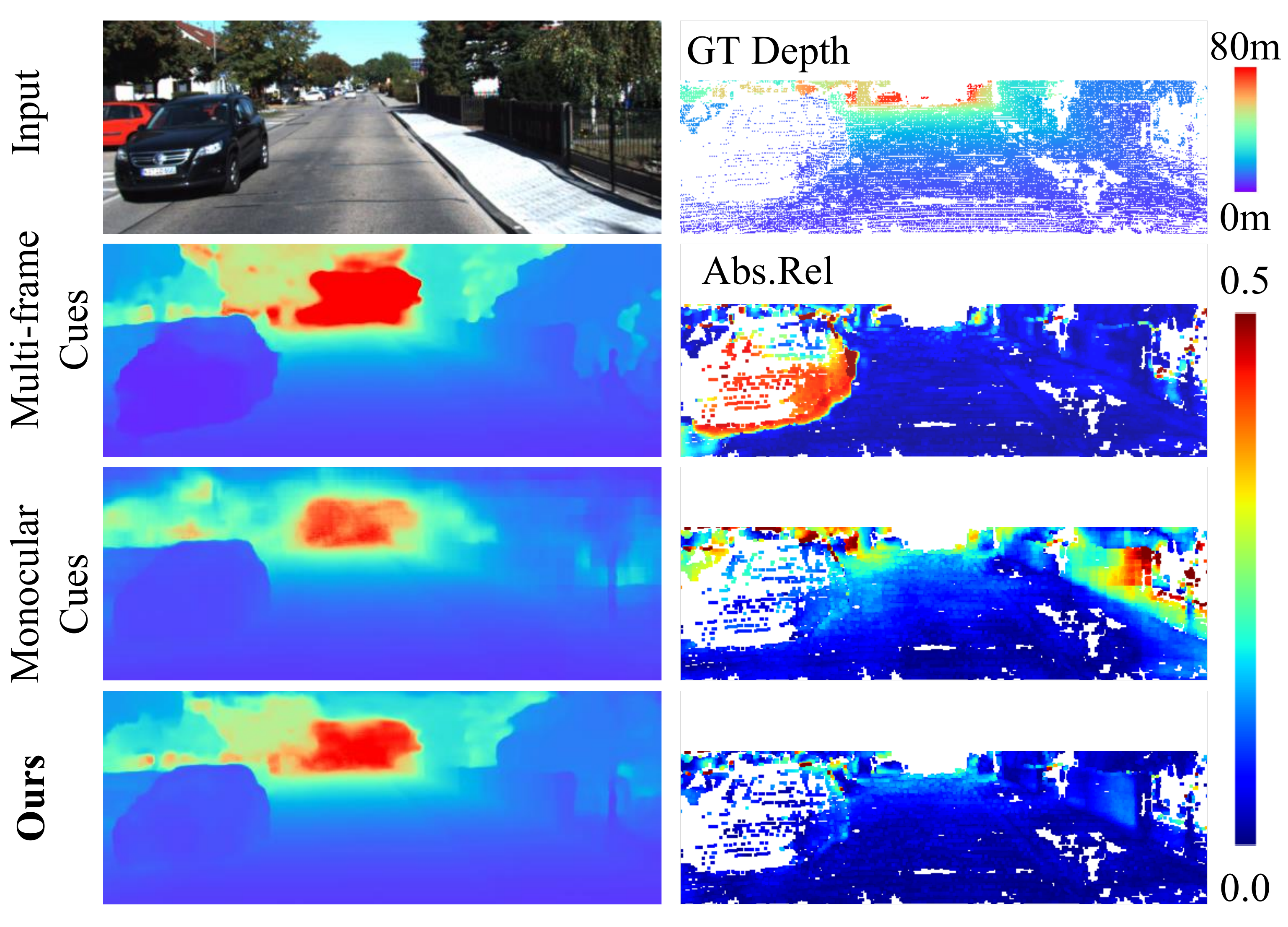}
\vspace{-12pt}
\caption{\textbf{Depth estimation in dynamic scenes.} Multi-frame predictions reserve high overall accuracy while degrading in dynamic areas. The monocular method better handles moving areas while suffering in static areas. Our method fuses both multi-frame and monocular depth cues for final prediction, yielding superior performance of the whole scene.}
\vspace{-18pt}
\label{fig:teaser}
\end{figure}

Although multi-frame methods are widely used in scene reconstruction \cite{yao2018mvsnet,gu2020cascade,wang2022mvster}, they encounter non-negligible challenges in dynamic scenes with dynamic areas (\eg, moving cars and pedestrians).
The dynamic areas cause corrupted values in the cost volume due to the violation of multi-view consistency \cite{feng2022disentangling,wang2022monocular} and mislead the network predictions. 
However, depth estimation for the dynamic areas is usually crucial in most applications \cite{geiger2012we,guizilini20203d,innmann2020nrmvs}. 
As shown in Fig. \ref{fig:teaser}, multi-frame depth estimation for the dynamic cars is more challenging than the static backgrounds.

\par
To handle the dynamic areas violating the multi-view consistency, a few multi-frame depth estimation methods \cite{wimbauer2021monorec,watson2021temporal,feng2022disentangling} try to identify and exclude the dynamic areas through an \emph{explicit} mask obtained relying on some assumptions or heuristics.
Specifically, some method \cite{wimbauer2021monorec} excludes the multi-frame cost volume relying on a learned dynamic mask and compensates the excluded areas with monocular features; 
some methods directly adjust the dynamic object locations in input images \cite{feng2022disentangling} or supervise multi-frame depth \cite{watson2021temporal} with predicted monocular depth.
However, these methods are usually sensitive to the explicit mask's quality, and the masks are obtained from additional networks or manually crafted criteria \cite{feng2022disentangling,wimbauer2021monorec,watson2021temporal}.
Despite better performances than pure multi-frame methods, these methods exhibit limited performance improvement compared with the additionally introduced \emph{monocular} cues (as shown in Tab. \ref{tab:improve_mono}), implying underutilized benefits from the \emph{fusion} of the multi-view and monocular cues. Although some self-supervised monocular depth estimation methods \cite{godard2019digging,casser2019depth,lee2021learning,cao2019learning, gordon2019depth,li2020unsupervised} also address the multi-view inconsistency issues, they mainly focus on handling the unfaithful self-supervision signals.

\par
To tackle the above issues, we propose a novel method that fuses the respective benefits from the monocular and multi-view cues, leading to significant improvement upon each individual source in dynamic scenes. 
We first analyze the behaviors of monocular and multi-frame cues in dynamic scenes, that the pure monocular method can generally learn good structural relations around the dynamic areas, and the {pure} multi-view cue preserves more accurate geometric properties in the static areas. We then unveil the effectiveness of leveraging the benefits of both depth cues by directly fusing depth volumes (Sec. \ref{sec:analysis}). Inspired by the above observations, beyond treating monocular cues as a local supplement of multi-frame methods \cite{wimbauer2021monorec,feng2022disentangling,watson2021temporal}, we propose a \emph{cross-cue fusion} (CCF) module to enhance the representations of multi-view and monocular cues with the other, and fuse them together for dynamic depth estimation. We use the spatially non-local relative intra-relations encoded in \emph{cross-observation attention} (CCA) weights from each source to guide the representation of the other, as shown in Fig. \ref{fig:cca}.
Specifically, the intra-relations of monocular cues can help to address multi-view inconsistency in dynamic areas, while the intra-relations from multi-view cues help to enhance the geometric property of the monocular representation, as visualized in Fig. \ref{fig:atten}. 
Unlike \cite{bae2022multi,wimbauer2021monorec,feng2022disentangling,watson2021temporal}, the proposed method unifies the input format of both cues as volumes and conducts fusion on them, which achieves better performances (as shown in Fig. \ref{fig:kitti}). 
The proposed fusion module is learnable and does not require any heuristic masks, leading to better generalization and flexibility. 

\par
Our main contributions are summarized as follows:
\begin{itemize}[topsep=-0.2cm, itemsep=-0.1cm]
\item We analyze multi-frame and monocular depth estimations in dynamic scenes and unveil their respective advantages in static and dynamic areas. Inspired by this, we propose a novel method that fuses depth volumes from each cue to achieve significant improvement upon individual estimations in dynamic scenes.

\item We propose a \emph{cross-cue fusion} (CCF) module that utilizes the \emph{cross-cue attention} to encode non-local intra-relations from one depth cue to guide the representation of the other. Different from methods using local masks, the attention weights learn mask-free global geometric information according to the geometric properties of each depth cue (as shown in Fig. \ref{fig:atten}).

\item The proposed method outperforms the state-of-the-art method in dynamic areas with a significant error reduction of $21.3\%$ while retaining its superiority in overall performance on KITTI. It also achieves the best generalization ability on the DDAD dataset in dynamic areas than the competing methods.
\end{itemize}

\begin{figure*}[t]
\centering
\vspace{-20pt}
\includegraphics[width=0.95\linewidth]{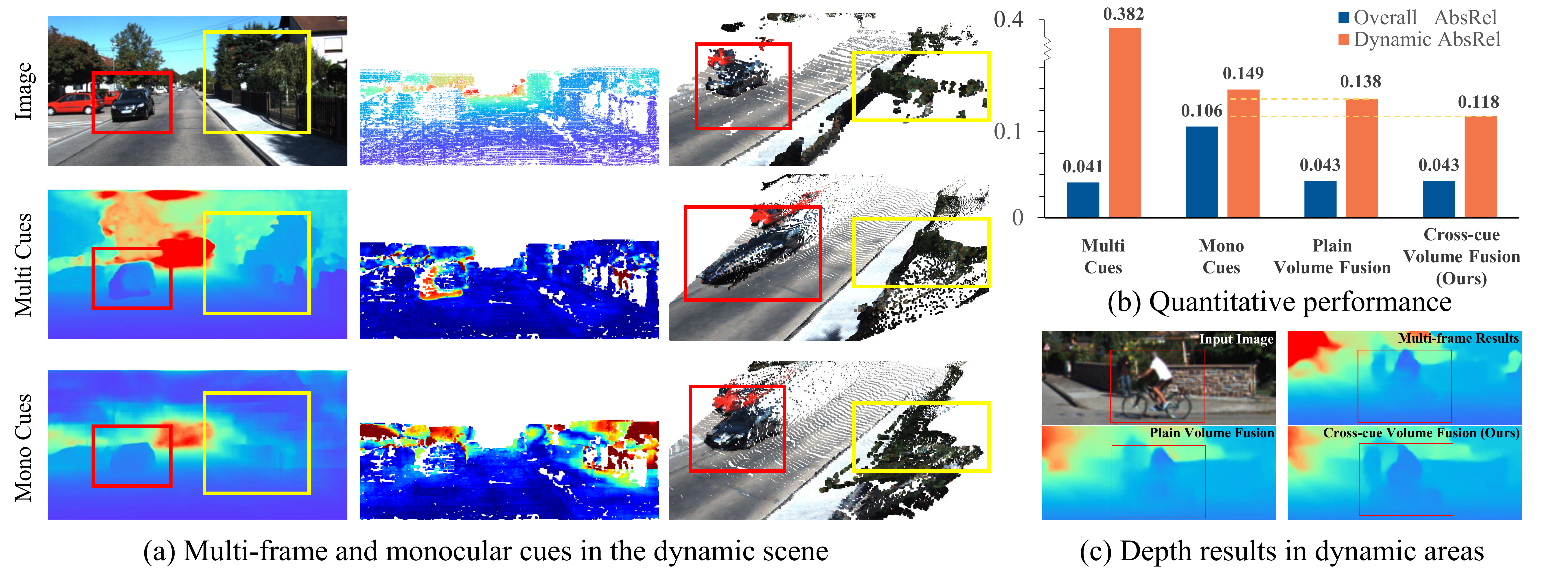}
\vspace{-10pt}
\caption{\textbf{Multi-frame and monocular cues in the dynamic scene.} a) Multi-frame cues preserve accurate geometric properties in the static area (yellow box), while the monocular cues learn good structural relations in dynamic areas (red box). b) Multi-frame and monocular cues show respective benefits in different areas. While our plain volume fusion scheme shows obvious performance improvement, the proposed \emph{cross-cue fusion} demonstrates better capabilities to handle dynamic depth. c) Depth predictions that show progressive improvements from the plain volume fusion and cross-cue volume fusion.}
\vspace{-15pt}
\label{fig:analysis}
\end{figure*}

\section{Related Work}\label{sec:related_work}
\noindent \textbf{Learning-based multi-frame depth estimation.}
Learning depth estimation from multiple images has attracted much attention these years. Current methods \cite{yao2018mvsnet,gu2020cascade,yang2022mvs2d,yao2019recurrent,duzceker2021deepvideomvs,wang2022itermvs} typically construct a cost volume using homography warping \cite{yao2018mvsnet} between multi-view images. 
Under the premise that the scene is static, the cost volume encodes the probabilities of different depth hypotheses for each pixel, which can be regularized by 3D CNNs to yield final depth prediction. 
Aiming at recovering the accurate structure of static scenes, many endeavors have been made in improving the quality of the cost volume \cite{gu2020cascade,mi2022generalized}, network efficiency \cite{yao2019recurrent,gu2020cascade, yang2022mvs2d} as well as the temporal consistency \cite{duzceker2021deepvideomvs,wang2022itermvs}, \textit{etc}. The effectiveness of static scene reconstruction has sparked several attempts \cite{watson2021temporal,guizilini2022multi,wimbauer2021monorec} that extend the multi-frame depth estimation to the large-scale outdoor scenes. The cost-volume between temporal consistent images provides extra depth information than image contexts \cite{watson2021temporal}. However, for the dynamic areas that violate the static scene assumption, the cost volume provides wrong depth probabilities, which mislead the network to produce wrong depth even under supervised learning \cite{wimbauer2021monorec}.
In this regard, processing the dynamic areas has become one of the main challenges for multi-frame depth estimation in outdoor scenes.  

\noindent \textbf{Dynamic depth estimation in outdoor scenes.}
Dynamic areas are ubiquitous in real-world scenarios and are important in applications such as autonomous driving.
Some literature \cite{casser2019depth,lee2021learning, li2020unsupervised, cao2019learning, gordon2019depth} seeks to handle dynamic objects in self-supervised monocular depth estimation, where the network inputs a single image and the dynamic areas mainly affect the supervisory signals. They typically identify possible moving objects by semantic \cite{klingner2020self} or instance segmentation \cite{casser2019depth,lee2021learning,cao2019learning,brickwedde2019mono}, then conduct robust learning by masking out dynamic areas during loss computation \cite{klingner2020self} or directly model the object motion \cite{casser2019depth,cao2019learning,gordon2019depth, brickwedde2019mono}. However, these methods differ from the topic discussed in this paper in that: 1) we focus on addressing multi-frame dynamic depth estimation issue, where the main challenge lies in the corrupted cost volumes than the self-supervised loss; 2) our method is supervised so that the robustness of loss function is beyond the main concern of this task.

\par
In the context of multi-frame depth estimation, current methods leverage the depth information from the single image to improve depth in dynamic areas. Manydepth \cite{watson2021temporal} proposes a self-discovered mask and supervises the potential dynamic areas with monocular depth. MonoRec \cite{wimbauer2021monorec} proposes a motion segmentation network to mask out the dynamic areas in the cost volume and use only monocular image features to infer depth.
Feng \etal \cite{feng2022disentangling} proposes to correct the dynamic object locations (with instance mask) in the image plane using monocular depth before computing the cost volume. Despite the higher dynamic results than pure multi-frame estimations, their performances are quite comparable \cite{feng2022disentangling,bae2022multi,watson2021temporal}, if not worse than, their proposed monocular branch (as shown in Tab. \ref{tab:improve_mono}). Moreover, they require pre-computed instance masks \cite{wimbauer2021monorec,feng2022disentangling} that bring extra computation burden for network training or inference.
There also exists another multi-frame method \cite{bae2022multi} which guides multi-frame cost reconstruction with a large monocular network. However, due to the reliance on monocular network predictions, it demonstrates weaker generalization capabilities than multi-frame methods (Tab. \ref{tab:ddad}). In contrast, our methods do not require any pre-computed object masks and achieve obvious improvement upon both monocular and multi-frame predictions in dynamic areas while retaining good generalization abilities across datasets.

\begin{figure*}[t]
\centering
\vspace{-15pt}
\includegraphics[width=0.95\textwidth]{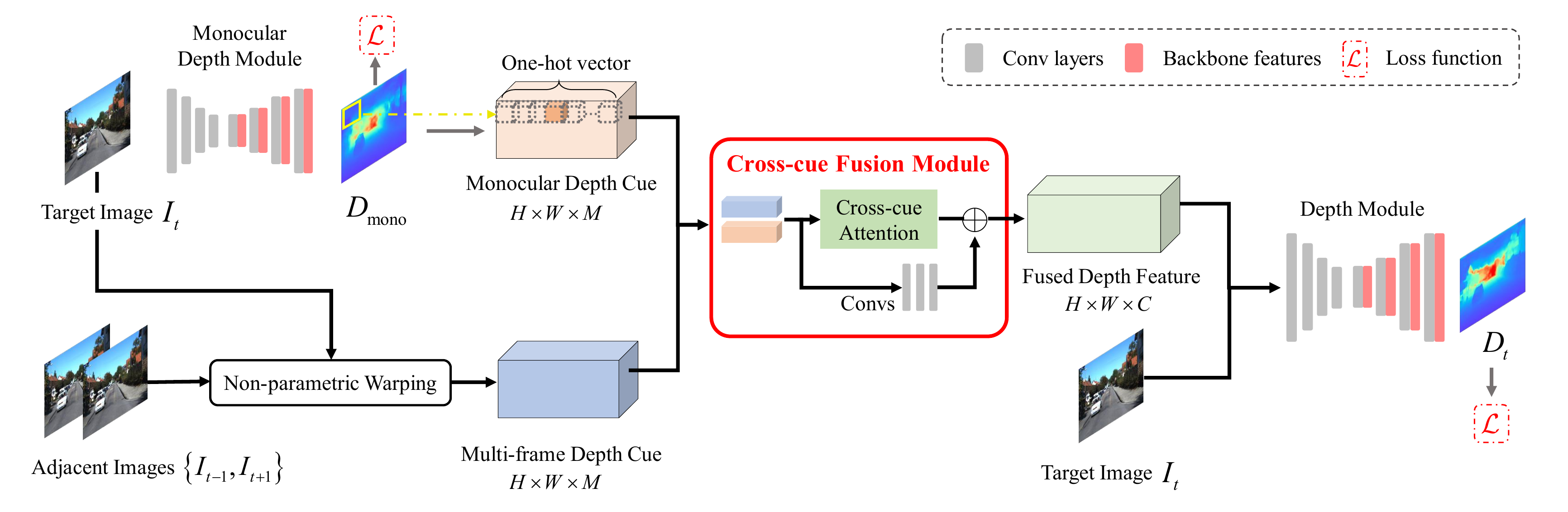}
\vspace{-12pt}
\caption{\textbf{Overview of the proposed method.} We first extract multi-frame depth cues with cost volume and monocular depth cues using one-hot depth volume. Then, we fuse the two volumes with the proposed cross-cue fusion module (CCF) to yield an improved fused depth feature. The fused depth feature is sent to the depth module for final depth estimation.}
\vspace{-10pt}
\label{fig:overview}
\end{figure*}

\section{Analyses on Multi-view \& Monocular Cues}
\label{sec:analysis}
Focusing on the dynamic scenes, we first analyze the behaviors of the depth estimation methods relying on multi-view and monocular cues, as in Sec. \ref{sec:ana_1}. Beyond the overall performance on the whole scene, we specifically study their behaviors in the dynamic area. Then we analyze how the proper fusion of multi-view and monocular cues may benefit depth estimation in dynamic areas in Sec. \ref{sec:ana_2}. 

\subsection{Behaviours of Multi-view and Monocular Cues} 
\label{sec:ana_1}
We implement two depth estimation methods with the two types of cues, \ie, Multi Cues and Mono Cues in Fig. \ref{fig:analysis}, and train them on KITTI \cite{geiger2012we}, where the U-Net \cite{wimbauer2021monorec} is used. Specifically, the model with multi-view cues (Multi Cues) takes the cost volume as the input. To analyze the performance in the dynamic areas, we use the dynamic mask from \cite{wimbauer2021monorec} computed by thresholding photometric error and depth inconsistency.
\par
As shown in Fig. \ref{fig:analysis} (b), multi-frame depth estimation generally achieves high overall accuracy relying on the multi-view clues. But the performance in the dynamic areas is degraded due to the violation of the multi-view consistency in the input cost volume. 
Fig. \ref{fig:analysis} (a) shows that the multi-frame method generates fine 3D structures in static areas (the yellow box), whereas its estimation of the moving car (the red box) is corrupted.
Compared with the multi-frame method, the monocular depth estimation method achieves worse overall performance as shown in Fig. \ref{fig:analysis} (b), without the geometric cues. On the other hand, it does not suffer from multi-view inconsistency and thus performs better in dynamic areas. In Fig. \ref{fig:analysis} (a), we can observe that the two depth cues show respective benefits in different areas and different aspects of the dynamic scene.

\subsection{Potential Benefits of Fusion}
\label{sec:ana_2}
To handle the dynamic areas violating the multi-view consistency, previous multi-frame depth estimation methods \cite{wimbauer2021monorec,feng2022disentangling,watson2021temporal} try to use the monocular information as the input for the dynamic areas. 
However, the improvement is limited due to the underutilized benefits of both cues, as discussed in Sec. \ref{sec:intro}. Especially the performance in the dynamic areas is easily bounded by the monocular cues. 
\par
We seek to explore better fusion schemes to use the complementary benefits of these two cues, as discussed in Sec. \ref{sec:ana_1}. 
Apart from using the monocular cues in the dynamic area, we also hope the geometric information learned with the multi-view cues in the static areas can be propagated to boost the monocular cues in dynamic areas, which requires a more comprehensive fusion process. 
\par
To unify the information from both multi-frame and monocular cues, we encode both depth cues in the form of a cost volume \cite{yao2018mvsnet} and a one-hot depth volume transferred from the estimated monocular depth map (in Sec. \ref{sec:depth_volume}). In the analyses, we first consider the \emph{plain volume fusion} using concatenation and convolution operations. This learnable fusion scheme brings obvious improvement in dynamic areas, implying proper fusion methods help to utilize the potential benefits in dynamic scenes. Based on the feasibility of learnable fusion, we further investigate the \emph{cross-cue volume fusion} and it exhibits further improvement (in Fig. \ref{fig:analysis} (b), (c)), which indicates a better way to utilize depth cues as introduced in the next section.

\section{The Proposed Method}
\label{sec:method}
We aim to learn depth $D_{t}$ of the target image $I_{t}$ from a short image sequence, with known or estimated camera parameters $K, T$. In this paper, we define the image sequence as $\{I_{t-1},I_{t},I_{t+1}\}$, where $\{I_{t-1}, I_{t+1}\}$ are adjacent images to the target frame $I_{t}$, $K, T$ are camera intrinsic and extrinsic provided as known values. The goal is to conduct accurate estimations of the dynamic scenes that contain various challenging dynamic objects.
\subsection{Overview}\label{sec:overview}
As shown in Fig. \ref{fig:overview}, the proposed method consists of three major parts - the multi-view and monocular cues are first represented as volumes through cost volume construction and depth one-hot vector transformation, the cross-cue fusion (CCF) module then fuses both multi-frame and monocular cues by leveraging attention mechanisms \cite{gong2021memory,vaswani2017attention,yan2022dual}, \ie, extracting relative intra-relations of each cue to guide the other to yield improved geometric representations of dynamic scene structure. The depth module takes the fused representation to estimate the final depth.
\par

\subsection{Representing Monocular and Multi-view Cues}
\label{sec:depth_volume}
\noindent\textbf{Multi-view cues as cost volume.} 
We represent the multi-view cues by computing the cost volume following the pipeline of multi-view stereo (MVS) \cite{yao2018mvsnet,wimbauer2021monorec}. We warp the adjacent images $\{I_{t-1}, I_{t+1}\}$ to the target view using $K, T$ and a set of depth hypotheses $d \in \{d_{k}\}_{k=1}^{M}$ uniformly sampled in the inverse depth space $[\frac{1}{d_\text{min}}, \frac{1}{d_\text{max}}]$, where $M$ denotes the number of depth hypotheses and is set to $32$ in our paper. We construct the multi-frame cost volume $C_\text{multi} \in \mathbb{R}^{H\times W \times M}$ by measuring the pixel-wise similarity between the warped images and the target image, using SSIM \cite{wang2004image} as introduced in \cite{wimbauer2021monorec}. For each pixel $(i,j)$ of $C_\text{multi} \in [0,1]^{H\times W \times M}$, channel positions $k\in\{1,\ldots, M\}$ with large matching scores indicates a higher possibility to include real scene depth. 

\par
\noindent\textbf{Monocular cues as depth volume.} 
We construct the monocular cues by estimating the single-view depth map and then transform the depth into one-hot depth volume. We leverage a U-Net architecture \cite{wimbauer2021monorec} without any complex designs for single-view depth estimation, yielding monocular prediction $ D_\text{mono} = f_{\theta}^\text{mono}(I_{t})$, where $D_{\text{mono}} \in \mathbb{R}^{H\times W}$ and $f_{\theta}^\text{mono}$ is the monocular network. To facilitate smooth fusion between the multi-frame and monocular cues, different from methods using single-view features \cite{wimbauer2021monorec,wang2022mvster} or depth values \cite{feng2022disentangling, watson2021temporal}, we transform the whole depth map $D_{\text{mono} }$ into the depth volume $C_\text{mono} \in \{0,1\}^{H\times W \times M}$ by converting each absolute depth value to a one-hot vector with
\begin{equation}
    C_{\text{mono},(i,j)}[k] = \{1 \hspace{1ex} \big| \hspace{1ex} d_\text{mono} \in (d_{k-1}, d_{k}] \}_{k=1}^{M},
\end{equation}
where $C_{\text{mono},(i,j)}\in \{0,1\}^M$ represents the one-hot vector in $C_{\text{mono}}$ corresponding to the pixel at $(i,j)$, $d_\text{mono}$ is the pixel-wise depth value. Note that we also try some soft representations \cite{bhat2021adabins,bae2022multi} but observe no obvious improvement.

\begin{figure}[t]
\centering
\vspace{-5pt}
\includegraphics[width=0.98\linewidth]{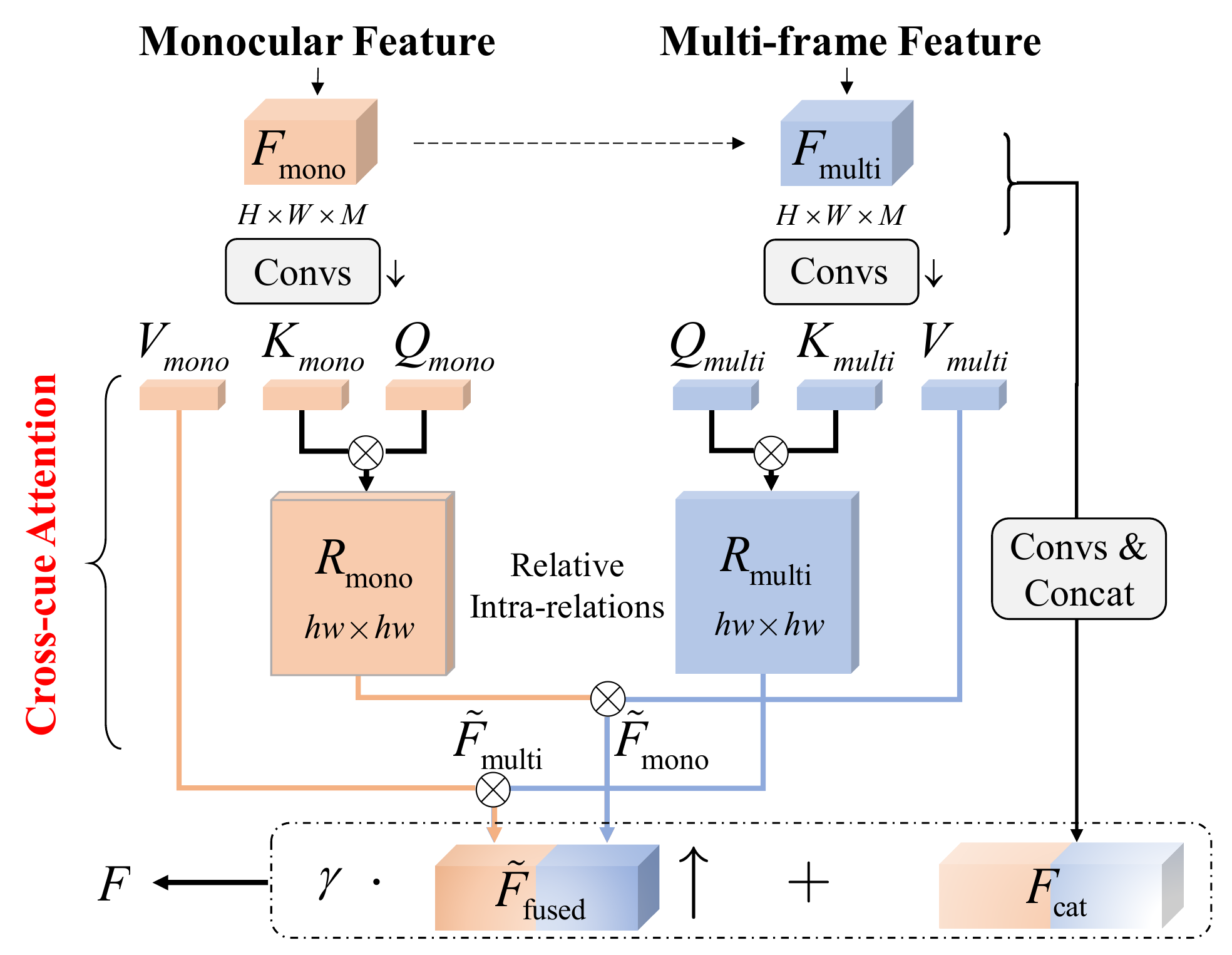}
\vspace{-5pt}
\caption{\textbf{The cross-cue fusion (CCF) module.} Taking both multi-frame and monocular depth volume as input, the CCF module enhances multi-frame depth features with the relative intra-relations of monocular depth volume ($R_\text{mono}$). Meanwhile, the intra-relations of the multi-frame depth volume ($R_\text{multi}$) also enhance the monocular module, yielding enhanced depth features for the final depth estimation.}
\vspace{-15pt}
\label{fig:cca}
\end{figure}

\subsection{The Cross-cue Fusion Module}
We propose a cross-cue fusion (CCF) module to fuse the multi-frame and monocular depth cues. Different from methods \cite{watson2021temporal,feng2022disentangling,wimbauer2021monorec} that utilize the masked local depth cues for dynamic depth estimation, the CCF module fuses the whole depth cues using depth volumes, retaining the potential to leverage both benefits for further improvement.

\par
\noindent \textbf{The Cross-cue fusion pipeline}. 
Given volumes $C_\text{multi}, C_\text{mono} \in \mathbb{R}^{H \times W \times M}$, we first process them via shallow convolution layers, yielding down-sampled monocular and multi-frame depth features $F_\text{multi}, F_\text{mono}$ in shape $\mathbb{R}^{h \times w \times M}$. We then feed $F_\text{multi}$ and $F_\text{mono}$ to the proposed \textit{cross-cue attention} (CCA) to enhance each depth cue by extracting the relative intra-relations from the other

\begin{equation}
\begin{aligned}
    \widetilde{F}_\text{multi} = {\text{CCA}_\text{multi}}(F_\text{mono}, F_\text{multi}), \\
    \widetilde{F}_\text{mono} = {\text{CCA}_\text{mono}}(F_\text{multi}, F_\text{mono}),
\end{aligned}
\end{equation}
the enhanced features are concatenated to yield the fused feature $\widetilde{F}_\text{fused}$. To retain detailed information from initial depth cues, we process the input depth cues via $F_\text{cat} = \text{Cat}(\text{Conv}(C_\text{multi}), \text{Conv}(C_\text{mono}))$ and add the residual connection. The final cross-cue features can be written as
\begin{equation}\label{eq:residual}
    F = \gamma \widetilde{F}_\text{fused} \uparrow + F_\text{cat},
\end{equation}
where $\gamma$ is a learned weighting factor and `$\uparrow$' denotes the up-sampling operation. The fused feature $F$ is then sent to the depth network along with the image context features to yield the final depth prediction $D_{t} \in \mathbb{R}^{H \times W}$.

\begin{figure}[t]
\centering
\vspace{-5pt}
\subfloat[Input image]{
 \begin{minipage}[c]{0.32\linewidth}
          \centering
          \includegraphics[width=1\linewidth]{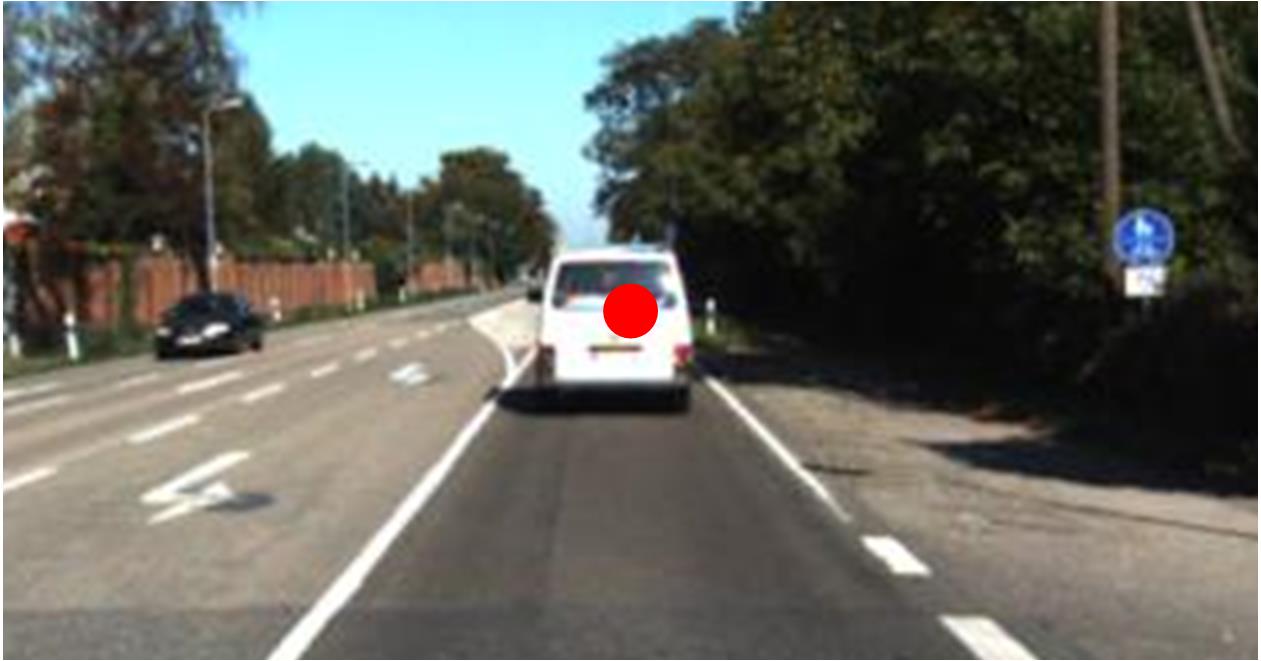}\\
        \includegraphics[width=1\linewidth]{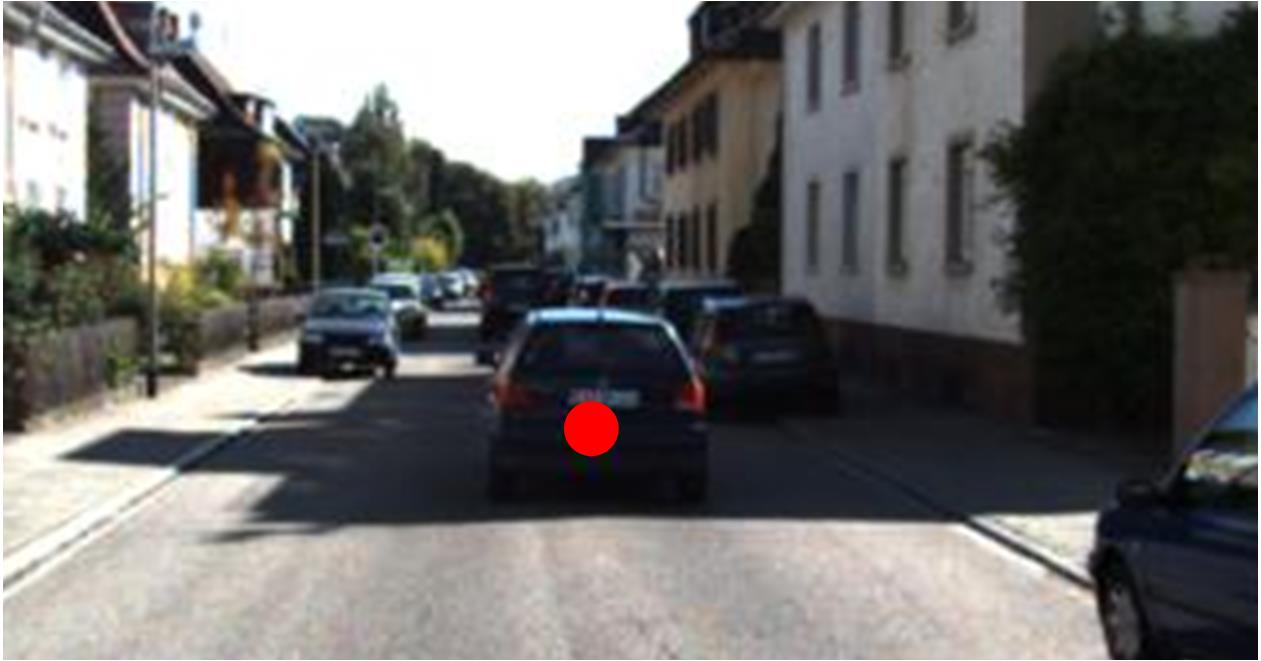}
      \end{minipage} \hspace{-5pt}
} 
\subfloat[$R_\text{mono}$ atten. map]{
 \begin{minipage}[c]{0.32\linewidth}
    \centering
    \includegraphics[width=1\linewidth]{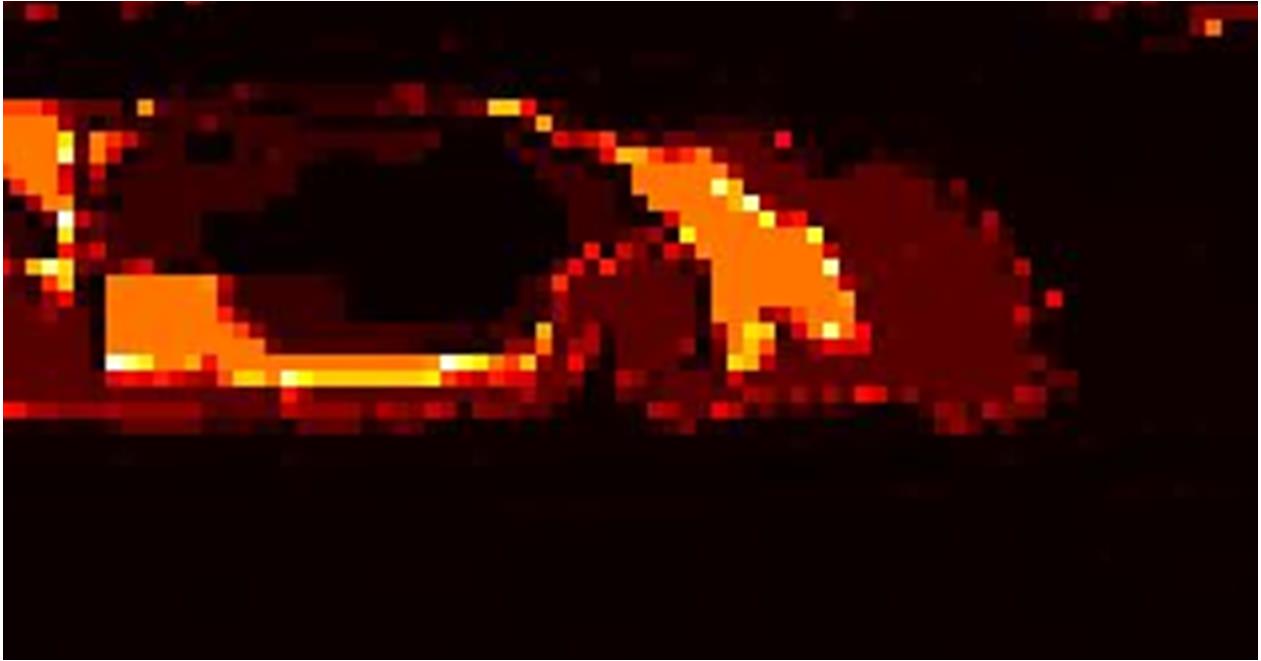}
    \\
    \includegraphics[width=1\linewidth]{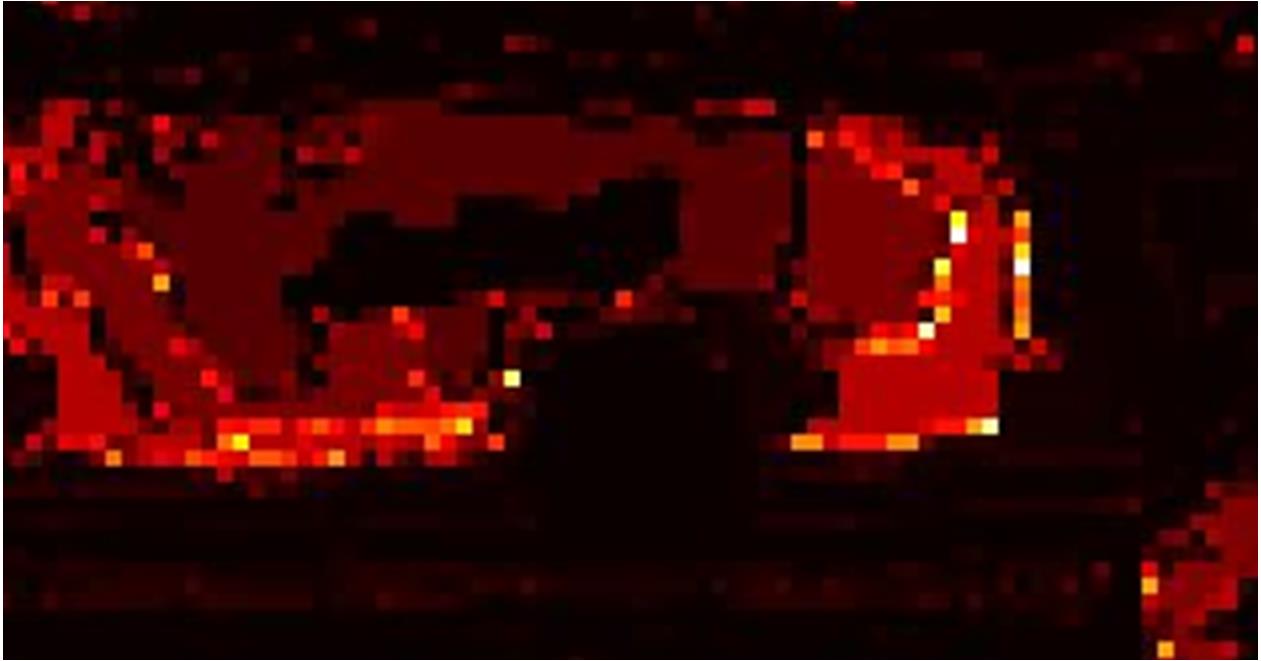}
     \end{minipage} \hspace{-5pt}
} 
\subfloat[$R_\text{multi}$ atten. map]{
 \begin{minipage}[c]{0.32\linewidth}
    \centering        \includegraphics[width=1\linewidth]{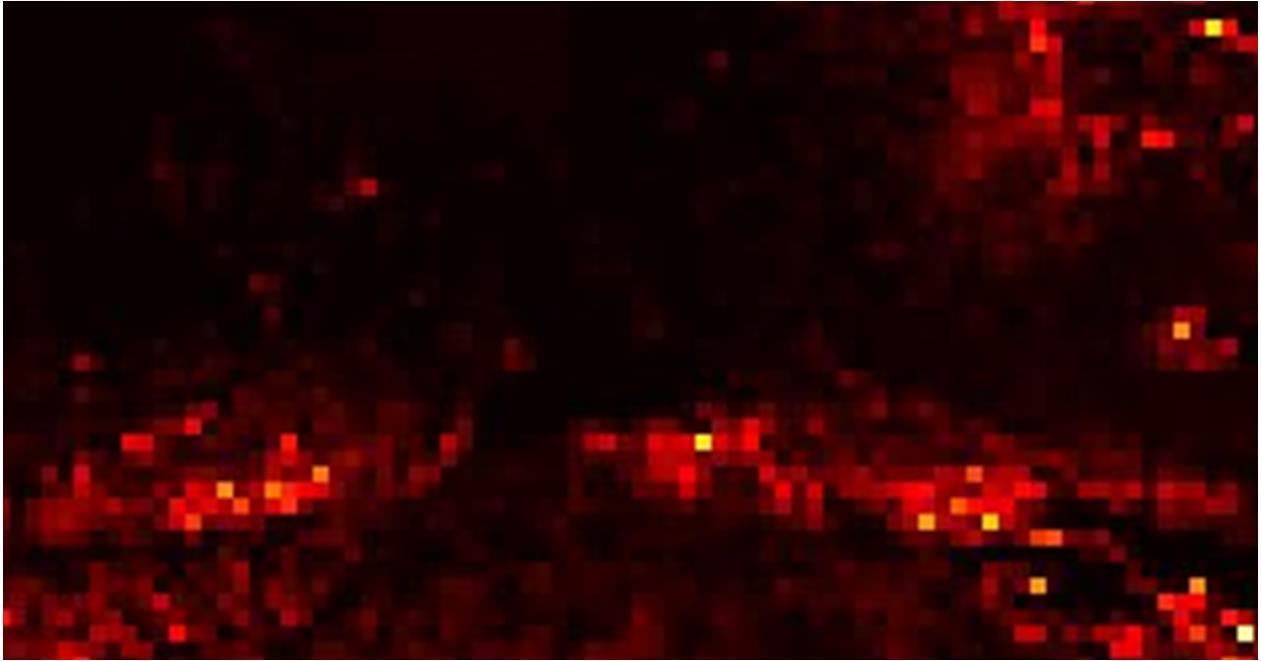}
    \\
     \includegraphics[width=1\linewidth]{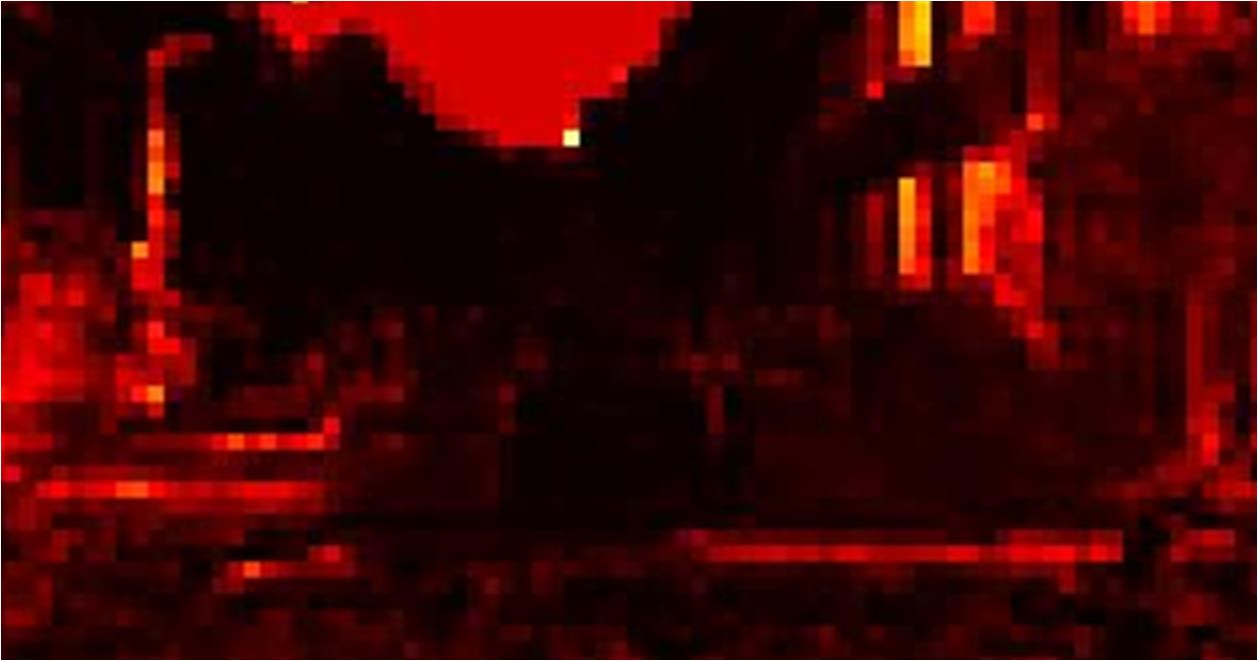}
     \end{minipage}
}
\centering
\vspace{-5pt}
\caption{\cam{\textbf{Attention maps of the dynamic position (red dots) examples}. $R_\text{mono}$ boosts multi-frame features by attending monocular features around the dynamic areas for better use of the monocular cues in the corresponding areas. $R_\text{multi}$ enhances monocular features by attending multi-frame features in the static areas which have more reliable geometric information.}}
\vspace{-10pt}
\label{fig:atten}
\end{figure}

\renewcommand\arraystretch{1.15}
\begin{table*}[ht]
\centering
\vspace{-10pt}
\scalebox{0.85}{
\begin{tabular}{c|p{3cm}<{\raggedright}|c|c|c|c|c|c|c|c|p{1.45cm}<{\centering}|p{1.45cm}<{\centering}}
\hline
Eval  &  Method                     & Back.                        &  Reso.     & Sup.       & \cellcolor{my_blue}Abs Rel & \cellcolor{my_blue}Sq Rel & \cellcolor{my_blue}RMSE  & \cellcolor{my_blue}RMSE$_{log}$ & \cellcolor{my_red}$\delta < 1.25$ &\cellcolor{my_red} $\delta < 1.25^{2}$ & \cellcolor{my_red} $\delta < 1.25^{3}$  \\ \hline

\multirow{8}{*}{\rotatebox{90}{\makebox{Overall}}} & Manydepth \cite{watson2021temporal}        & Res-18 &   MR     & \cellcolor{my_self}M  & 0.071          & 0.343       &3.184          & 0.108          & 0.945          & 0.991         &       0.998    \\

& DynamicDepth\cite{feng2022disentangling}       & Res-18 & MR & \cellcolor{my_self}M  & 0.068          & 0.296         & 3.067          & 0.106          & 0.945        & 0.991          & 0.998          \\

& MonoRec \cite{wimbauer2021monorec}  & Res-18 & MR & \cellcolor{my_semi}D$^{\ast}$  & 0.050          & 0.290         &2.266          & 0.082          & 0.972          & 0.991          &0.996          \\
&\textbf{Ours}        & Res-18 & MR & \cellcolor{my_sup}D  & \textbf{0.043}               &    \textbf{0.151}           &   \textbf{2.113}        &       \textbf{0.073}        &       \textbf{0.975}         &         \textbf{0.996}      &          \textbf{0.999}      \\
& MaGNet \cite{bae2022multi}  & Effi-B5 & MR & \cellcolor{my_sup}D  & 0.057          & 0.215         &2.597          & 0.088          & 0.967          & \textbf{0.996}          & \textbf{0.999}         \\

&\textbf{Ours}        & Effi-B5 & MR & \cellcolor{my_sup}D  & {0.046}               &    {0.155}           &   {2.112}        &       {0.076}        &       {0.973}         &         \textbf{0.996}      &          \textbf{0.999}      \\

\cline{2-12}

& MaGNet \cite{bae2022multi}  & Effi-B5 & HR & \cellcolor{my_sup}D  & 0.043          & 0.135         & 2.047          & 0.082          & 0.981          & \textbf{0.997}          & \textbf{0.999}         \\

&\textbf{Ours}        & Effi-B5 & HR &\cellcolor{my_sup}D  & \textbf{0.039}               &    \textbf{0.103}           &   \textbf{1.718}        &       \textbf{0.067}        &      \textbf{0.981}         &    \textbf{0.997}         &          \textbf{0.999}  \\
\hline \hline

\multirow{8}{*}{\rotatebox{90}{\makebox{Dynamic}}} & Manydepth \cite{watson2021temporal}        & Res-18 & MR     & \cellcolor{my_self}M  & 0.222          & 3.390       &7.921          & 0.237          & 0.676          & 0.902         &   0.964        \\

& DynamicDepth\cite{feng2022disentangling}       & Res-18 & MR & \cellcolor{my_self}M  & 0.208          & 2.757         & 7.362          & 0.227          & 0.682        & 0.911          & 0.971          \\

& MonoRec \cite{wimbauer2021monorec}  & Res-18 & MR & \cellcolor{my_semi}D$^{\ast}$  & 0.360          & 9.083         &10.963          & 0.346          & 0.590          & 0.882          &0.780          \\

&\textbf{Ours}        & Res-18 & MR & \cellcolor{my_sup}D  & {0.118}               &    {0.835}           &   {4.297}        &       {0.146}        &       {0.871}         &         {0.975}      &          {0.990}      \\ 

& MaGNet \cite{bae2022multi}  & Effi-B5 & MR & \cellcolor{my_sup}D  & 0.141          & 1.219         &4.877          & 0.168          & 0.830          & 0.955          & 0.986         \\

&\textbf{Ours}        & Effi-B5 & MR & \cellcolor{my_sup}D  & \textbf{0.111}               &    \textbf{0.768}           &   \textbf{4.117}        &       \textbf{0.135}        &       \textbf{0.881}         &         \textbf{0.980}      &          \textbf{0.994}      \\ 

\cline{2-12}
& MaGNet \cite{bae2022multi}  & Effi-B5 & HR & \cellcolor{my_sup}D  & 0.140          & 1.060         &4.581          & 0.202          & 0.834          & 0.954          & 0.982         \\

&\textbf{Ours}        & Effi-B5 & HR & \cellcolor{my_sup}D  & \textbf{0.112}               &    \textbf{0.830}           &   \textbf{4.101}        &       \textbf{0.137}        &       \textbf{0.885}         &         \textbf{0.978}      &         \textbf{0.992}      \\ 
\hline

\hline
\end{tabular}
}
\vspace{-5pt}
\caption{\textbf{Quantitative comparisons on KITTI \cite{geiger2012we} Odometry dataset}. `Back.' denotes the network backbone. `Reso.' denotes the image resolutions, where `MR' refers to the resolution of $256\times 512$ and `HR' is $352\times 1216$. In the `Sup.' column, \colorbox{my_self}{`M'} are self-supervised methods, \colorbox{my_semi}{`D$^{\ast}$'} refers to semi-supervised methods trained with pseudo GT depth, while \colorbox{my_sup}{`D'} denotes fully-supervised methods. Color \textcolor{my_blue}{\textbf{blue}} denotes `lower is better', while \textcolor{my_red}{\textbf{red}} means `higher is better'. The best results are in \textbf{bold}.}
\vspace{-15pt}
\label{tab:kitti_odom}
\end{table*}

\noindent \textbf{Cross-cue attention.} 
The cross-cue attention (CCA) targets utilizing the relative intra-relation of one depth cue to improve the geometric information of another. Since the CCA modules are deployed in a parallel manner as shown in Fig. \ref{fig:cca}, we introduce $\widetilde{F}_\text{multi} = {\text{CCA}_\text{multi}}(F_\text{mono}, F_\text{multi})$ in detail as an example.
\par
Given depth features $F_\text{mono}, F_\text{multi} \in \mathbb{R}^{h \times w \times M}$, we transform $F_\text{mono}$ into query feature $Q_\text{mono}$ and key feature $K_\text{mono}$, then transform $F_\text{multi}$ into value feature $V_\text{multi}$ using convolution operation $f(\cdot,\theta)$
\begin{equation}
\begin{aligned}
    Q_\text{mono} = f(F_\text{mono}, \theta^{Q}_\text{mono}), \\
    K_\text{mono} = f(F_\text{mono}, \theta^{K}_\text{mono}), \\
    V_\text{multi} = f(F_\text{multi}, \theta^{V}_\text{multi}),\\
\end{aligned}
\end{equation}
after reshaping all features into size $(hw,M)$, we compute the non-local relative intra-relations of monocular features by matrix multiplication $\otimes$ followed by softmax operation
\begin{equation}\label{eq:enhance}
R_\text{mono} = {\text{Softmax}}(Q_\text{mono} \otimes K_\text{mono}^{T}),
\end{equation}
where $R_\text{mono} \in \mathbb{R}^{hw \times hw}$ stands for the non-local intra-relations of the monocular depth cue. We then utilize $R_\text{mono}$ to improve the geometric representations of the multi-frame feature $V_\text{multi}$ by
\begin{equation}
\widetilde{F}_\text{multi} = R_\text{mono} \otimes V_\text{multi}, 
\end{equation}
where $\widetilde{F}_\text{multi}$ denotes the improved multi-frame representations benefited from monocular depth cues. Similar operations are done for $\widetilde{F}_\text{mono} = {\text{CCA}_\text{mono}}(F_\text{multi}, F_\text{mono})$, where $R_\text{multi}$ stands for the intra-relations of multi-frame cues that can be used to improve monocular depth feature $V_\text{mono}$. Please refer to Fig. \ref{fig:cca} for details.
\par

\noindent\textbf{Effectiveness of the CCA.} As shown in Fig. \ref{fig:atten}, despite using the same CCA operation, \cam{the intra-relations $R_\text{mono}$ and $R_\text{multi}$ attend different areas for improving the dynamic depth, showing their ability to capture respective benefits from monocular (better dynamic depth) and multi-frame cues (better static depth). This learned property enables unbounded dynamic depth performance upon both predictions, \ie, the monocular depth can be improved by attending static depth (using $R_\text{multi}$) from multi-frame cues, and the improved monocular depth in dynamic areas will further be propagated (with $R_\text{mono}$) to multi-frame predictions.
}

\subsection{Loss Function}
Given the predicted depth $D$ and the ground truth depth $\hat{D}$, the loss can be described as 
\begin{equation}
    \mathcal{L}(D,\hat{D}) = \beta \mathcal{L}_\text{SI}(D,\hat{D}) + \mathcal{L}_\text{VNL}(D,\hat{D}),
\end{equation}
where $\mathcal{L}_\text{SI}$ denotes the scale-invariant loss \cite{eigen2014depth,bhat2021adabins}, $\mathcal{L}_\text{VNL}$ is the virtual normal loss \cite{yin2021virtual,yin2019enforcing}, $\beta$ is the weighing factor which is set to $4$. Since we have both monocular depth $D_\text{mono}$ and the final depth prediction $D_\text{t}$, the final loss $\mathcal{L}_\text{final}$ is
\begin{equation}
    \mathcal{L}_\text{final} = \mathcal{L}(D_\text{mono},\hat{D}) + \mathcal{L}(D_{t},\hat{D}). 
\end{equation}

\section{Experiments}\label{sec:experiments}
In this section, we compare our method with state-of-the-art multi-frame depth estimation methods \cite{watson2021temporal,feng2022disentangling,wimbauer2021monorec,bae2022multi} for the dynamic scenes, including self-supervised methods \cite{watson2021temporal,feng2022disentangling}, semi-supervised method \cite{wimbauer2021monorec} as well as fully supervised method \cite{deng2009imagenet} (Sec. \ref{sec:kitt_res}). We then conduct ablation studies (Sec. \ref{sec:ablation}) to evaluate different variants of our method. We also validate the generalization ability (Sec. \ref{sec:ddad}) and evaluate different methods' improvements upon monocular networks for depth estimation in dynamic areas (Sec. \ref{sec:improve_mono}).

\renewcommand\arraystretch{1.15}
\begin{table*}[h]
\small
\center
\vspace{-5pt}
\scalebox{0.95}{
\begin{tabular}{c|c|p{5.5cm}<{\raggedright}|ccccc|c}
\hline
\multirow{2}{*}{\#} & \multirow{2}{*}{Category}    & \multirow{2}{*}{Variant}  & \multicolumn{5}{c|}{Dynamic}                                                                                                                       & Overall \\ \cline{5-9} 
                    &                              &                       & \multicolumn{1}{c|}{\cellcolor{my_blue}Abs Rel} & \multicolumn{1}{c|}{\cellcolor{my_blue}Abs Sq} & \multicolumn{1}{c|}{\cellcolor{my_blue}RMSE} & \multicolumn{1}{c|}{\cellcolor{my_blue}RMSE$_{log}$} & \cellcolor{my_red}$\delta \le 1.25$ & \cellcolor{my_blue} AbsRel  \\ \hline
1                   & \multirow{2}{*}{\makecell[c]{Depth with  \\ single cues}}    & Pure multi-frame cues                 & \multicolumn{1}{c|}{0.382}        & \multicolumn{1}{c|}{7.167 }       & \multicolumn{1}{c|}{10.292}     & \multicolumn{1}{c|}{0.35}             &         0.509              &    0.041     \\ \cline{1-1}

2                   &                              & Pure monocular cues                & \multicolumn{1}{c|}{0.149}        & \multicolumn{1}{c|}{1.369}       & \multicolumn{1}{c|}{5.282}     & \multicolumn{1}{c|}{0.178}             &           0.810            &      0.106   \\ \hline

3                   & \multirow{2}{*}{\makecell[c]{Volume fusion \\ with masks}} & Self-discovered mask \cite{watson2021temporal}            & \multicolumn{1}{c|}{0.130}        & \multicolumn{1}{c|}{0.990}       & \multicolumn{1}{c|}{4.692}     & \multicolumn{1}{c|}{0.160}             &         0.837              &       0.043  \\ \cline{1-1}

4                   &                              & MaskNetwork \cite{wimbauer2021monorec}       & \multicolumn{1}{c|}{0.220}        & \multicolumn{1}{c|}{2.896}       & \multicolumn{1}{c|}{6.299}     & \multicolumn{1}{c|}{0.223}             &   0.735                    &   0.040      \\ \cline{1-1}

\hline
5                   & \multirow{9}{*}{\makecell[c]{Volume fusion \\ without masks}} & Stack \& 3D Convs               & \multicolumn{1}{c|}{0.154}        & \multicolumn{1}{c|}{1.479}       & \multicolumn{1}{c|}{5.866}     & \multicolumn{1}{c|}{0.189}             &           0.777            &    0.046     \\ \cline{1-1}

6                   &                              &Stack \& 3D U-Net \cite{gu2020cascade}                   & \multicolumn{1}{c|}{0.155}        & \multicolumn{1}{c|}{1.444}       & \multicolumn{1}{c|}{5.762}     & \multicolumn{1}{c|}{0.191}             &       0.772                &     \textbf{0.040}    \\ \cline{1-1}
7                   &                             & Concat \& 2D Convs              & \multicolumn{1}{c|}{0.138}        & \multicolumn{1}{c|}{1.124}       & \multicolumn{1}{c|}{5.110}     & \multicolumn{1}{c|}{0.174}             &        0.815               &      0.043   \\  \cline{1-1} \cline{3-9}

 \cline{3-9} 

8                  &                              & \textbf{Ours} CCF w./o. $R_\text{multi}$          & \multicolumn{1}{c|}{0.124}        & \multicolumn{1}{c|}{0.939}       & \multicolumn{1}{c|}{4.610}     & \multicolumn{1}{c|}{0.154}             &      0.855                &    0.043     \\ \cline{1-1}

9                  &                              & \textbf{Ours} CCF w./o. $R_\text{mono}$                  & \multicolumn{1}{c|}{0.123}        & \multicolumn{1}{c|}{0.926}       & \multicolumn{1}{c|}{4.545}     & \multicolumn{1}{c|}{0.153}             &    0.861                   &    0.043     \\ \cline{1-1}

10                  &                              & \textbf{Ours} CCF w./ only intra-cue self-attention           & \multicolumn{1}{c|}{0.122}        & \multicolumn{1}{c|}{0.896}       & \multicolumn{1}{c|}{4.544}     & \multicolumn{1}{c|}{0.152}             &             0.860          &    0.042     \\ \cline{1-1}

11                  &                              & \textbf{Ours} CCF w./o. residual connection              & \multicolumn{1}{c|}{0.130}        & \multicolumn{1}{c|}{0.961}       & \multicolumn{1}{c|}{4.616}     & \multicolumn{1}{c|}{0.157}             &             0.840          &    0.048     \\ \cline{1-1}

12                  &                              & \cam{\textbf{Ours} depth module w./o.} $I_{t}$         & \multicolumn{1}{c|}{{0.126}}        & \multicolumn{1}{c|}{{0.954}}       & \multicolumn{1}{c|}{{4.636}}     & \multicolumn{1}{c|}{{0.155}}             &      {0.844}                 &    0.042     \\ \cline{1-1} \cline{3-9}

13                  &                              & \textbf{Ours} CCF - full         & \multicolumn{1}{c|}{\textbf{0.118}}        & \multicolumn{1}{c|}{\textbf{0.835}}       & \multicolumn{1}{c|}{\textbf{4.297}}     & \multicolumn{1}{c|}{\textbf{0.146}}             &      \textbf{0.871}                 &    0.043     \\ \hline
\end{tabular}
}
\vspace{-5pt}
\caption{\textbf{Ablation experiments on KITTI}. We show the results of different fusion types of multi-frame and monocular cues. `CCF' denotes the proposed cross-cue fusion module. `Dynamic' denotes dynamic depth errors while `Overall' refers to the overall depth error.}
\vspace{-15pt}
\label{tab:ablation}
\end{table*}

\subsection{Datasets}
\noindent\textbf{KITTI.} \hspace{2pt} We follow \cite{wimbauer2021monorec} to evaluate dynamic sequential data on KITTI Odometry dataset, which contains 13666 training and 8634 testing samples. We use the dynamic masks provided by \cite{wimbauer2021monorec} for evaluation. More than 1300 samples in the test set contain dynamic objects.  We evaluate both medium-resolution (MR: $256\times 512$) and high-resolution (HR: $352\times 1216$) depth results, and the metrics are computed within the 0-80m range.
\par
\noindent\textbf{DDAD.} \hspace{2pt} We use the forward-facing cameras with 3848 testing samples to evaluate the methods' generalization ability. Since there is no pre-defined dynamic mask for DDAD, we construct it by warping adjacent images with GT depth and selecting instance masks with high photometric error. More than 70\% of the test set contains dynamic objects. All methods are evaluated within the 0-80m depth range.

\subsection{Implementation Details}
We implement our method with Pytorch \cite{paszke2017automatic} and train it using NVIDIA TITAN RTX GPUs. Unless specified, both the monocular network and the depth network are consistent with the depth module of \cite{wimbauer2021monorec}
, with ResNet-18 \cite{he2016deep} as backbone using ImageNet \cite{deng2009imagenet} pre-trained parameters. We train our model for 80 epochs using the Adam \cite{kingma2014adam} optimizer and learning rate of $10^{-4}$, which further drops to $10^{-5}$ after 65 epochs. The batch size is set to 8.

\subsection{Results on KITTI}\label{sec:kitt_res}
\par
\begin{figure*}[t]
\centering
\vspace{-12pt}
\subfloat{
    \begin{minipage}[c]{0.01\linewidth}
        \centering
        \scriptsize
        \rotatebox{90}{\makebox{Input\&GT}}
    \end{minipage}
    \begin{minipage}[c]{0.95\linewidth}
        \centering
        \includegraphics[width=1\linewidth]{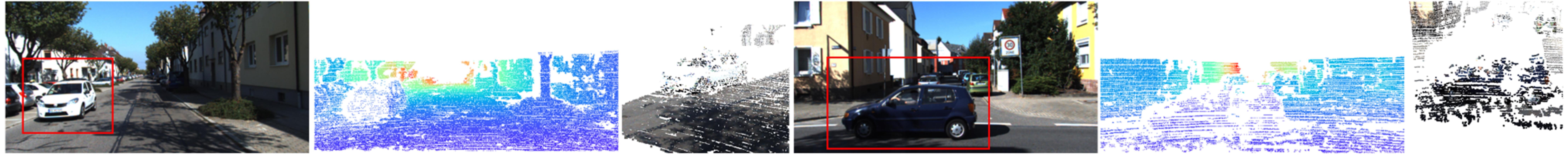}
    \end{minipage}
} \vspace{0.5pt}\\
\subfloat{
    \begin{minipage}[c]{0.01\linewidth}
        \centering
        \scriptsize
        \rotatebox{90}{\makebox{ManyD\cite{watson2021temporal}}}
    \end{minipage}
    \begin{minipage}[c]{0.95\linewidth}
        \centering
        \includegraphics[width=1\linewidth]{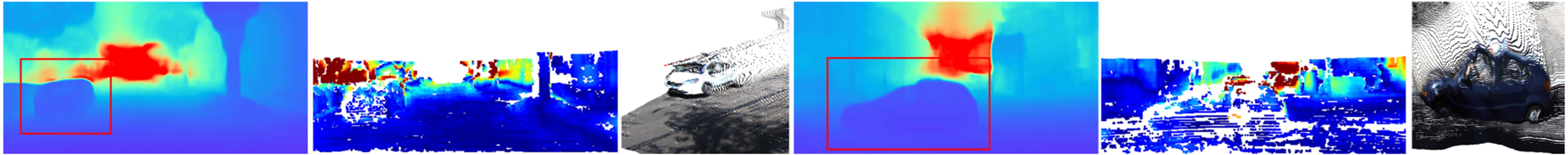}
    \end{minipage}
} \vspace{0.5pt}\\
\subfloat{
    \begin{minipage}[c]{0.01\linewidth}
        \centering
        \scriptsize
        \rotatebox{90}{\makebox{DyDepth\cite{feng2022disentangling}}}
    \end{minipage}
    \begin{minipage}[c]{0.95\linewidth}
        \centering
        \includegraphics[width=1\linewidth]{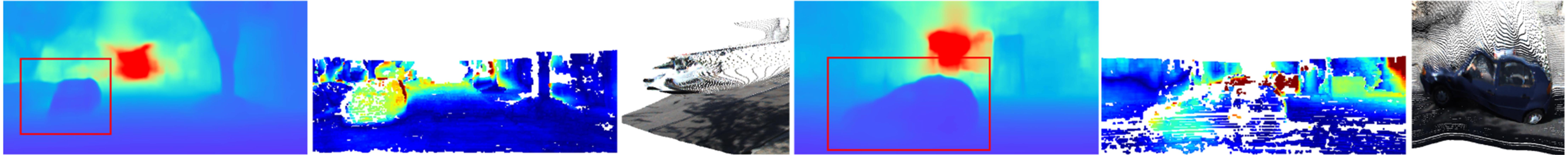}
    \end{minipage}
} \vspace{0.5pt}\\
\subfloat{
    \begin{minipage}[c]{0.01\linewidth}
        \centering
        \scriptsize
        \rotatebox{90}{\makebox{MonoRec\cite{wimbauer2021monorec}}}
    \end{minipage}
    \begin{minipage}[c]{0.95\linewidth}
        \centering
        \includegraphics[width=1\linewidth]{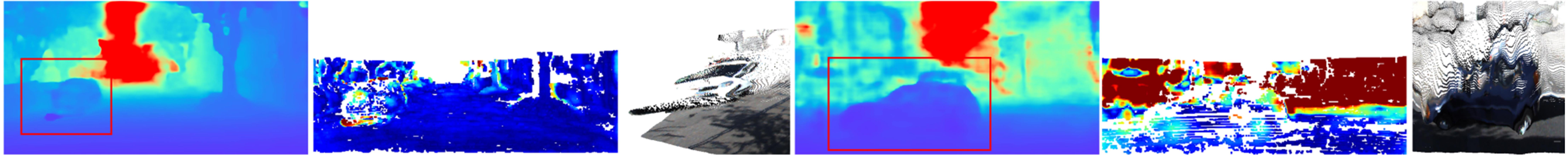}
    \end{minipage}
} \vspace{0.5pt}\\
\subfloat{
    \begin{minipage}[c]{0.01\linewidth}
        \centering
        \scriptsize
        \rotatebox{90}{\makebox{MaGNet\cite{bae2022multi}}}
    \end{minipage}
    \begin{minipage}[c]{0.95\linewidth}
        \centering
        \includegraphics[width=1\linewidth]{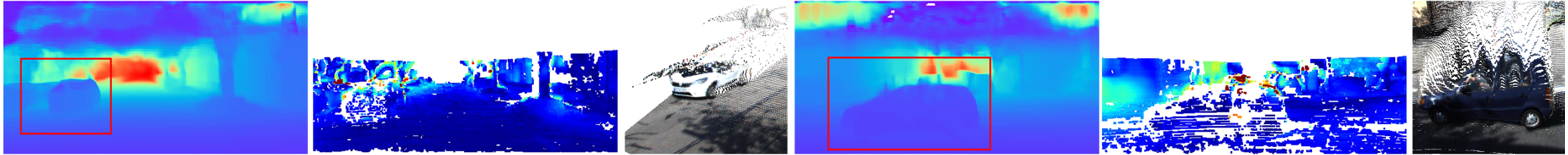}
    \end{minipage}
}\vspace{0.5pt} \\
\subfloat{
    \begin{minipage}[c]{0.01\linewidth}
        \centering
        \scriptsize
        \rotatebox{90}{\makebox{\textbf{Ours}}}
    \end{minipage}
    \begin{minipage}[c]{0.95\linewidth}
        \centering
        \includegraphics[width=1\linewidth]{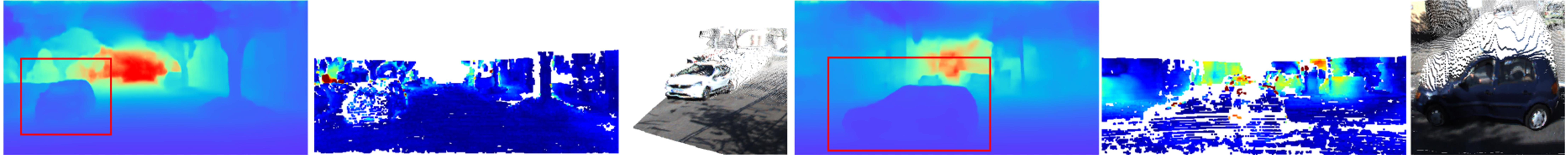}
    \end{minipage}
}
\vspace{-8pt}
\caption{\textbf{Qualitative results on KITTI dataset \cite{geiger2012we}.} From left to right: depth predictions (dynamic objects are highlighted with red boxes), error maps, and the reconstructed point clouds of dynamic areas. Our method achieves the best dynamic results and reconstructs more reasonable object shapes than state-of-the-art methods.}
\vspace{-6pt}
\label{fig:kitti}
\end{figure*}

We show both overall and dynamic performances of different methods in Table \ref{tab:kitti_odom}. Our method achieves the best performance in dynamic depth estimation. Specifically, it outperforms the SOTA fully-supervised MaGNet \cite{bae2022multi} with $21.28\%$ reduction of Abs.Rel (0.141$\rightarrow$0.111), using the same Efficient-B5 \cite{tan2019efficientnet} backbone. Significant improvements are also observed compared to self/semi-supervised methods \cite{watson2021temporal,feng2022disentangling,wimbauer2021monorec}. 
Besides the obvious improvement in dynamic areas, our method also outperforms other methods in most metrics for overall performance. 
Qualitative results are shown in Fig. \ref{fig:kitti}. While the dynamic areas generally lead to performance decline for other multi-frame methods, our method conducts obviously better estimations in the moving objects, while retaining the overall accuracy.

\renewcommand\arraystretch{1.25}
\begin{table*}[ht]
\centering
\scalebox{0.85}{
\begin{tabular}{c|p{3.5cm}<{\raggedright}|c|c|c|c|c|c|p{1.45cm}<{\centering}|p{1.45cm}<{\centering}}
\hline
Eval  &  Method                     & Backbone                       & \cellcolor{my_blue}Abs Rel & \cellcolor{my_blue}Sq Rel & \cellcolor{my_blue}RMSE  & \cellcolor{my_blue}RMSE$_{log}$ & \cellcolor{my_red}$\delta < 1.25$ &\cellcolor{my_red} $\delta < 1.25^{2}$ & \cellcolor{my_red} $\delta < 1.25^{3}$  \\ \hline

\multirow{3}{*}{\rotatebox{90}{\makebox{Overall}}} 
& MonoRec \cite{wimbauer2021monorec}  & Res-18 & \textbf{0.158}          & 3.102         &\textbf{7.553}          & \textbf{0.227}          & \textbf{0.854}          & \textbf{0.931}          &\textbf{0.961}          \\
& MaGNet \cite{bae2022multi}  & Effi-B5  & 0.208          &\underline{2.641}        &    10.739      &   0.382        &    0.620       & 0.878          &0.942         \\
&\textbf{Ours}        & Res-18    & \textbf{0.158}               &    \textbf{2.416}           &   \underline{9.855}        &       \underline{0.299}        &       \underline{0.747}         &    \underline{0.894}         &          \underline{0.947}      \\ 
\hline

\multirow{3}{*}{\rotatebox{90}{\makebox{Dynamic}}} 
& MonoRec \cite{wimbauer2021monorec}  & Res-18   & 0.544          & 16.703         &16.116          & 0.482          & 0.460          & 0.667          &0.798          \\
& MaGNet \cite{bae2022multi}  & Effi-B5  & \underline{0.266}          & \underline{3.982}         &\underline{11.715}          & \underline{0.398}          & \underline{0.462}          & \underline{0.815}          &\underline{0.917}       \\
&\textbf{Ours}        & Res-18  & \textbf{0.234}               &    \textbf{3.611}           &   \textbf{11.007}        &       \textbf{0.331}        &       \textbf{0.576}         &         \textbf{0.835}      &          \textbf{0.921}      \\
\hline
\end{tabular}
}
\vspace{-5pt}
\caption{\textbf{Generalizations on DDAD \cite{guizilini20203d} dataset}. The best results are in \textbf{bold} and the second best results are \underline{underlined}. Our method achieves a competitive overall performance with other methods while achieving the best result in dynamic areas.}
\vspace{-10pt}
\label{tab:ddad}
\end{table*}

\subsection{Ablation Study}\label{sec:ablation}
As shown in Sec. \ref{sec:analysis}, the way to fuse the two volumes in our method influences the final dynamic depth performance. 
As shown in Tab. \ref{tab:ablation}, we fuse the multi-frame and monocular volumes in ways that the explicit mask is needed (`Volume fusion with masks') and the explicit mask is not needed (`Volume fusion without masks'). We also evaluate the variants of our method (`\textbf{Ours} CCF') as individual mask-free methods.
Row \#1\textasciitilde2 shows individual depth results leveraging pure multi-frame and monocular cues. 

\par
\noindent\textbf{Volume fusion with explicit masks.} 
We leverage the computed \cite{watson2021temporal} or learned masks \cite{wimbauer2021monorec} to fuse our volumes from two depth cues. As shown in row \#3\textasciitilde4, The dynamic mask generated by \cite{wimbauer2021monorec} does not surpass the baseline fusion method. While the self-discovered mask \cite{watson2021temporal} shows certain improvements in dynamic areas, it is computed using heuristics and thus has uncontrolled dynamic mask quality, which leads to more restricted performances than ours. 

\par
\noindent\textbf{Volume fusion without masks.} Besides our method that uses a mask-free fusion scheme, we also fuse the volumes without any mask using convolutions. We first stack the depth volumes in a new dimension and process the fused features with layers of 3D convolutions as well as the 3D U-Net, which is commonly used in the MVS methods \cite{yao2018mvsnet,gu2020cascade}. As shown in row \# 5\textasciitilde 6 of Table \ref{tab:ablation}, the 3D convolutions have no obvious improvement in dynamic areas. However, as we concatenate the two depth volumes and process them with 2D convolution layers (as in Fig. \ref{fig:analysis}), we observe further improvements (row \#7) upon the pure monocular results, but this mask-free fusion scheme still lags behind our method with obvious margins.

\par
\noindent\textbf{Design choices of the \cam{network} modules.} We evaluate different variants of our method as an individual type of mask-free fusion method.
We respectively remove the multi-frame (row \#8) and monocular (row \#9) intra-relations and directly feed the other cues to the fused feature. Then, we implement another variant of our method with intra-cue self-attention in the proposed CCF (row \#10). We also disable the residual connection of $F_\text{cat}$ (row \#11) and the \cam{input target image $I_{t}$ of the depth module (row \#12)}. Results show the effectiveness and necessity of the proposed technical designs, which achieves the best performance.

\begin{table}[]
\centering
\scalebox{0.90}{
\begin{tabular}{l|c|c|c}
\hline
Method       & Mono. Err. & Final Err. & Err. Redu. \\ \hline
Manydepth \cite{watson2021temporal}   &    0.212       &   0.222    &      $-4.72\%$       \\
Dynamicdepth \cite{feng2022disentangling} &   0.214        &  0.208    &    2.83\%         \\
MaGNet \cite{bae2022multi}      &      0.153     &    0.141   &      7.84\%       \\ \hline
\textbf{Ours} - Res.18         &      0.149     &  0.118  &   \textbf{20.81\%}      \\ 
\textbf{Ours} - Res.50         &      0.145     &  0.116  &   \textbf{20.00\%}      \\ 

\hline
\end{tabular}}
\vspace{-5pt}
\caption{\textbf{Error reduction upon monocular estimation in dynamic areas}. We show different methods' \emph{dynamic} Abs.Rel depth errors (Final Err.) and their reduction (Err. Redu.) compared to their individual monocular performances (Mono. Err.). Our method achieves the \cam{largest error reduction over others and exhibits a consistent error reduction when a better backbone is used}.}
\vspace{-15pt}
\label{tab:improve_mono}
\end{table}

\subsection{Generalization on DDAD} \label{sec:ddad}
To validate the generalization ability of our method, we test the KITTI models of all supervised methods \cite{bae2022multi,wimbauer2021monorec} on the challenging DDAD dataset. As shown in Table \ref{tab:ddad}, previous methods either degrade on the dynamic areas due to the violation of multi-view consistency (\eg, AbsRel 0.544 for \cite{wimbauer2021monorec}) or exhibit worse overall performance owing to the cross-domain issue of the monocular network (\eg, AbsRel 0.208 for \cite{bae2022multi}). Our method shows more promising results that it not only outperforms other methods in dynamic depth estimation but also retains the advantage of multi-frame estimation with competitive overall performance.

\subsection{Improvement upon Monocular Estimation} \label{sec:improve_mono}
Though previous methods handle dynamic areas with monocular cues, the improvement is usually constrained by monocular results. In Tab. \ref{tab:improve_mono}, we show the final depth results in dynamic areas and compare them against the monocular results of each method. Our method achieves the \cam{largest Abs.Rel reduction} of $20.8\%$ compared to other methods which explicitly use the monocular network. Meanwhile, when using a backbone with larger capacities (ResNet-50), our method \cam{achieves consistent error reduction} upon the monocular network, showing its scalability and flexibility.

\section{Conclusion}\label{sec:conclusion}
We improve multi-frame dynamic depth estimation by fusing the multi-view and monocular depth cues with the proposed cross-cue fusion. Experiments show its effectiveness as well as the generalization ability. \textbf{Limitation}: \cam{there still exists a performance gap to fill, especially in dynamic areas, with a better fusion of the cues.
Meanwhile, the challenging occlusion issues for general multi-frame methods require further exploration.}
\par
{\noindent \textbf{Acknowledgements}
Y. Zhu, J. Sun, and Y. Zhang acknowledge the support from National Engineering Laboratory for Integrated Aero-Space-Ground-Ocean Big Data Application Technology. This work was partially supported by NSFC (No.U19B2037) and the Natural Science Basic Research Program of Shaanxi (No.2021JCW-03) to Y. Zhang; an ARC DECRA Fellowship DE230101591 to D. Gong.}

{\small
\bibliographystyle{ieee_fullname}
\bibliography{egbib}
}

\end{document}